\documentclass{article}

    \PassOptionsToPackage{numbers, compress}{natbib}

    \usepackage[preprint]{neurips_2024}

\usepackage{amsmath}
\usepackage{tabularray}
\usepackage{algorithm}
\usepackage{algpseudocode}
\usepackage[pdftex]{graphicx}
\usepackage[utf8]{inputenc} 
\usepackage[T1]{fontenc}    
\usepackage{hyperref}       
\usepackage{url}            
\usepackage{booktabs}       
\usepackage{amsfonts}       
\usepackage{nicefrac}       
\usepackage{microtype}      
\usepackage{xcolor}         
\usepackage{multicol}
\usepackage[inline]{enumitem}

\usepackage{listings}
\usepackage{wrapfig}
\usepackage{acronym}

\acrodef{IndepViews}[IndepViews]{Independent View Policy}
\acrodef{CoViews}[CoViews]{Cooperative View Policy}
\acrodef{SSL}[SSL]{Self-supervised learning}

\definecolor{codegreen}{rgb}{0,0.6,0}
\definecolor{codegray}{rgb}{0.5,0.5,0.5}
\definecolor{codepurple}{rgb}{0.58,0,0.82}
\definecolor{backcolour}{rgb}{0.95,0.95,0.92}

\lstdefinestyle{mystyle}{
    backgroundcolor=\color{backcolour},   
    commentstyle=\color{codegreen},
    keywordstyle=\color{magenta},
    numberstyle=\tiny\color{codegray},
    stringstyle=\color{codepurple},
    basicstyle=\ttfamily\footnotesize,
    breakatwhitespace=false,         
    breaklines=true,
    captionpos=b,                    
    keepspaces=true,                 
    numbers=left,                    
    numbersep=5pt,                  
    showspaces=false,                
    showstringspaces=false,
    showtabs=false,                  
    tabsize=2
}
\title{CoViews: Adaptive Augmentation Using Cooperative Views for Enhanced Contrastive Learning}

\author{%
  Nazim Bendib \\
  Higher National School of Computer Science (ESI ex INI) \\
  Algiers, Algeria\\
  \texttt{jn\_bendib@esi.dz} \\
}

\begin{document}

\maketitle

\begin{abstract}
\label{section:abstract}

Data augmentation plays a critical role in generating high-quality positive and negative pairs necessary for effective contrastive learning. 
However, common practices involve using a single augmentation policy repeatedly to generate multiple views, potentially leading to inefficient training pairs due to a lack of cooperation between views.
Furthermore, to find the optimal set of augmentations, many existing methods require extensive supervised evaluation, overlooking the evolving nature of the model that may require different augmentations throughout the training.
Other approaches train differentiable augmentation generators, thus limiting the use of non-differentiable transformation functions from the literature. 
In this paper, we address these challenges by proposing a framework for learning efficient adaptive data augmentation policies for contrastive learning with minimal computational overhead. Our approach continuously generates new data augmentation policies during training and produces effective positives/negatives without any supervision. 
Within this framework, we present two methods: 
\ac{IndepViews}, which generates augmentation policies used across all views, and 
\ac{CoViews}, which generates dependent augmentation policies for each view. This enables us to learn dependencies between the transformations applied to each view and ensures that the augmentation strategies applied to different views complement each other, leading to more meaningful and discriminative representations.
Through extensive experimentation on multiple datasets and contrastive learning frameworks, we demonstrate that our method consistently outperforms baseline solutions and that training with a view-dependent augmentation policy outperforms training with an independent policy shared across views, showcasing its effectiveness in enhancing contrastive learning performance.

\end{abstract}

\section{Introduction}
\label{section:introduction}

\ac{SSL} is a type of unsupervised learning that leverages the intrinsic structure of data to learn meaningful representations without explicit supervision, surpassing even representations learned through supervised methods in an increasing number of cases \cite{simclr, moco}. Among \ac{SSL} methods, contrastive learning emerges as a highly effective approach where the objective is to differentiate between positive pairs coming from the same data sample and negative pairs coming from different data samples. These pairs are often generated through data augmentation, with augmented views of the same sample serving as positive pairs and views of different samples as negative pairs.
Data augmentation has proved to be a crucial setup to enable contrastive representation learning, leading researchers to perform extensive supervised evaluations \cite{simclr, mocov2} to find the best augmentation policy for training. 
This policy is referred to as the “sweet spot” \cite{infomin} because it strikes a balance between distorting images and preserving task-relevant information. 
In addition to time and resource consumption for finding such policies, current contrastive learning methods \cite{simclr, moco, byol, simsiam} typically employ a fixed augmentation policy to generate challenging training pairs for an evolving encoder overlooking the fact that the difficulty of pairs depends on the encoder's evolving performance during training, thus suggesting the need for adaptive augmentation strategies. 
Moreover, utilizing a single augmentation policy to repeatedly generate views can be inefficient due to the lack of coordination between the generated views. This can result in cases where the views are relatively easy to distinguish, such as applying identical transformations to both views, potentially leading to a less sample-efficient training process.

In this work, we introduce an adaptive data augmentation technique that builds adaptive augmentation policies during training. This approach eliminates the need to train the model multiple times to evaluate different policies, reducing computational overhead. Our solution does not rely on auxiliary tasks \cite{faa, rotation} or differentiable data augmentation transformations \cite{viewmaker, infomin}. Instead, it can use standard data augmentation techniques commonly found in the literature \cite{autoaugment}, simplifying implementation and ensuring compatibility with existing frameworks.
Additionally, we introduce a variant of our solution where instead of learning a shared augmentation policy for all views and independently sampling transformations for each one of them, it learns conditional augmentation policies for each view. This implies that the transformations applied to one view depend on those applied to the other view. By establishing this interdependence, we ensure that all generated pairs are challenging, thereby enhancing sample efficiency. This cooperation between the views allows us to create difficult pairs for the encoder without heavily relying on highly distorting transformations.

\paragraph{Contribution summary.} We summarize our contributions as follows:
\begin{itemize}
    \setlength\itemsep{0pt}
    \item We propose a new framework for learning adaptive augmentation policies during training. This framework utilizes a novel reward function called Bounded InfoNCE, which enables us to evaluate good policies without the need for auxiliary tasks.
    \item We introduce two methods: 
    \begin{enumerate*}[label=(\arabic*)]
        \item \ac{IndepViews} for generating policies that are independently used to generate both views, and
        \item \ac{CoViews} for generating dependent cooperative views for more efficient training pairs.
    \end{enumerate*}
    \item Our method consistently outperforms baseline solutions across multiple datasets in terms of linear evaluation.
    \item We inspect and demonstrate through extensive experiments that learning and using dependent cooperative views through \ac{CoViews} results in better performance compared to using independent view policies from \ac{IndepViews}.
\end{itemize}

\section{Related work}

\subsection{Contrastive Learning}
Contrastive learning has emerged as a powerful paradigm in representation learning. By training a model to differentiate between positive and negative pairs of data instances, contrastive learning aims to learn meaningful representations that capture the inherent similarities and differences within the data. Through the use of carefully designed loss functions and augmentation techniques, contrastive learning has shown remarkable success across various tasks, notably in computer vision. One of the most famous frameworks in contrastive learning is SimCLR \cite{simclr}, which employs augmented views to generate positive/negative samples, where positive samples are augmented views of the same image and negative samples are augmented views of different images. Moco \cite{moco}, another successful approach, leverages a momentum-updated memory bank of old negative representations to enable the use of large batches of negative samples. Other methods emerged that rely solely on positive pairs to learn robust representations, such as BYOL \cite{byol} which employs an online encoder and a momentum encoder and learns discriminative representations by minimizing the similarity between the online predicted embedding and the momentum embedding. SimSiam \cite{simsiam} extends this approach by eliminating the momentum key encoder and using stop-gradient to address collapsing issues.

\subsection{Learning data augmentation policies}

Due to the huge impact of data augmentation \cite{dataaug_survey} in computer vision, many works \cite{autoaugment, randaugment, faa, population_augmentation, adversarial_autoaugment, gans_sl} have been dedicated to identifying optimal data augmentation strategies in supervised learning setups to enhance training efficiency and improve learned representations.
While automatic data augmentation has been extensively explored within supervised learning frameworks, its application within self-supervised setups has remained relatively underexplored, despite the critical role that data augmentation plays in contrastive learning. 
SelfAugment \cite{saa} introduced variants for RandAugment \cite{randaugment} and FastAutoAugment \cite{faa} to be used in a contrastive learning setup with no access to labels, called SelfRandAugment and SelfAugment, where they replaced the supervised evaluation of the augmentation policies with a self-supervised evaluation. They empirically demonstrated a strong correlation between linear probe accuracy and image rotation prediction \cite{rotation}, thus utilizing the latter for policy evaluation. 
Similarly, InfoMIN \cite{infomin} adversarially trained a differentiable flow-based network to map natural colors into new color channels, which were later split to generate the views. 
Another approach, Viewmaker \cite{viewmaker}, adversarially trained a fully differentiable network that generates perturbations to add to an input image to generate the views.  

In our work, we also aim to address the main drawbacks of these methods: SelfAugment requires multiple training iterations to build the augmentation policy before training the final model, whereas InfoMIN and Viewmaker use a differentiable network to generate the views in a fully differentiable adversarial training, limiting the usage of standard augmentation techniques and applicability to other modalities of data.


\section{Problem Formulation}
The goal of contrastive representation learning is to learn an embedding space in which similar sample pairs stay close to each other while dissimilar ones are far apart. The most commonly used contrastive loss function is the InfoNCE loss \cite{infonce}, where, in practice, 
given a batch of samples $X$, we generate two views $X_1 = \mathcal{T}_1(X)$ and $X_2 = \mathcal{T}_2(X)$ to form a positive pair, and pairs of views coming from different samples form negative pairs - with $\mathcal{T}_1$ and $\mathcal{T}_2$ being stochastic data augmentation functions. The InfoNCE loss function can be written as follows:

\begin{equation}
    \mathcal{L}_{NCE}(X_1,X_2)=\frac{1}{2N} \sum_{i=1}^{N} \mathcal{L}(X_{1,i}, X_{2,i}) + \mathcal{L}(X_{2,i}, X_{1,i})
\label{eq:infoNCE_my_formulation}
\end{equation}
with $\mathcal{L}(x_1,x_2)$ defined as
\begin{equation}
    \mathcal{L}(x_1,x_2)=-log \frac{e^{sim(f(x_1), f(x_2))/\tau}}{\sum_{x_k \in X_1 \cup X_2, x_1 \neq x_k}e^{sim(f(x_1), f(x_k))/\tau}}
\end{equation}
where $sim(\cdot,\cdot)$ is the cosine similarity and $\tau$ is the temperature parameter.
The goal is to learn a parametric encoder function $f$ that maximizes the similarity between the representations of positive pairs and minimizes it between the representations of negative pairs.  
Finally, the objective of contrastive learning can be written as the following optimization problem:
\begin{equation}
    \min_{f}\mathcal{L}_{NCE} \big( f(\mathcal{T}_1(X)),f(\mathcal{T}_2(X)) \big)
\label{eq:min_objective}
\end{equation}

Hard data augmentation has proved to be a crucial setup in contrastive learning framework \cite{simclr, moco, strong_augmentation}, though simply focusing on very hard augmentations distorts the image structure \cite{strong_augmentation}, resulting in difficult retrieval, so finding the best augmentation policy is still a tedious task.
In this work, We believe that optimal data augmentation transformations should be able to fulfill two conditions:
\begin{enumerate*}[label=(\arabic*)]
    \item They should be sufficiently challenging to decrease the mutual information between views, thereby enabling the model to learn discriminative representations;
    \item They should not be excessively challenging to the point that they discard all task-relevant information.
\end{enumerate*}
These conditions align with the InfoMIN \cite{infomin} principle, which claims that there exists a “sweet spot” where the mutual information (MI) between views is neither too high nor too low. 
One formulation for this problem can be to employ an adversarial training strategy similar to GANs \cite{gan}. Where we train the encoder $f$ to minimize the InfoNCE loss while using data augmentations to maximize the InfoNCE loss. Formally, the problem can be expressed as:
\begin{equation}
    \min_{f} \max_{g_1,g_2} \mathcal{L}_{NCE} \big( f(g_1(X)),f(g_2(X)) \big)
\label{eq:min_max_formulation}
\end{equation}
Where $g_1, g_2$ are transformation functions that generate views. 
While this approach may effectively learn data augmentation strategies verifying condition (1), it will certainly violate condition (2) by learning degenerate transformations that eliminate all task-relevant information from the views (e.g. generating overly dark or bright views) leading the whole training to fail. To ensure that the learned augmentation policy verifies condition (2), a sort of regularization is necessary to prevent degenerate solutions. 
InfoMIN \cite{infomin} achieves this by employing a flow-based model \cite{glow} as an invertible transformation function that maps natural colors to novel color spaces. The invertibility of such a function preserves task-relevant information, verifying condition (2). 
Viewmaker \cite{viewmaker} uses L1 regularization to limit the amount of perturbation to apply to the images.
Similarly, SelfAugment \cite{saa} adopts a linear image rotation prediction task to regularize the transformations. This is accomplished by learning an augmentation policy that minimizes the rotation cross-entropy loss of the generated views while simultaneously maximizing the InfoNCE loss.
These last methods apply the same augmentation policy to independently generate the views.
In this work, we aim to find augmentation policies that verify both conditions (1) and (2) without requiring additional implementation complexities while being able to learn a dependency between the augmentations generating the views, leading to better cooperation. 
Specifically, we aim to avoid the necessity for differentiable augmentation transformations, auxiliary tasks, or any further contrastive training.

\section{Adaptive Augmentation Policy}

In this section, we introduce our approach to continuously learning adaptive augmentation policies. Our method revolves around iteratively using the current encoder to search for challenging augmentation policies to improve its performance in subsequent epochs. See Appendix \ref{pseudo-algorithm} for a pseudo-algorithm. Our algorithm comprises three main phases:

\paragraph{Warmup phase}
We begin by training the encoder $f$ for a few warm-up epochs using random augmentations. This initial phase is crucial in our algorithm, as it establishes a baseline performance for the encoder. This baseline performance provides a foundation that enables us to later search for improved augmentation policies effectively. Without these warm-up epochs, the encoder lacks the context to discern whether pairs are difficult or not, as it is initialized randomly. 

The next two phases are executed iteratively until the end of the training:

\paragraph{Adaptive Augmentation Policy Generation phase}
In this phase, we utilize the current encoder $f$ to search for a new augmentation policy that challenges its current capabilities, by finding which augmentations is the encoder weak against and trying to focus on them in subsequent epochs.
More details about the search algorithm and the evaluation function are provided in \ref{section:policy_network} and \ref{section:bounded_infonce} respectively. 
To ensure smooth training, we don't rely solely on the new augmentation policy for contrastive training. Instead, we maintain a history of the $Q$ most recently learned policies. 
This is accomplished by assigning decreasing probabilities $( p_1 > p_2 > ... > p_Q )$ to each policy with $p_1$ being the probability of the most recent policy. This way, we first sample a policy from the queue of the $Q$ most recent policies and then we sample data augmentation transformations from that chosen policy.
We run this phase each K epoch to reduce the computational overhead.

\paragraph{Contrastive Training phase}
During this phase, we employ the learned augmentation policies to train the encoder using contrastive loss. No additional modifications are introduced within the contrastive training loop. We consider this aspect a significant advantage of our method, as it allows for a seamless integration of adaptive augmentation policies into any contrastive learning algorithm without requiring any modifications to the latter (see appendix \ref{appendix:moco} for experiments using MoCo \cite{moco}).

\subsection{Adaptive Data Augmentation Policy Network}
\label{section:policy_network}



\begin{figure}[t]
    \centering
    \includegraphics[width=1\linewidth]{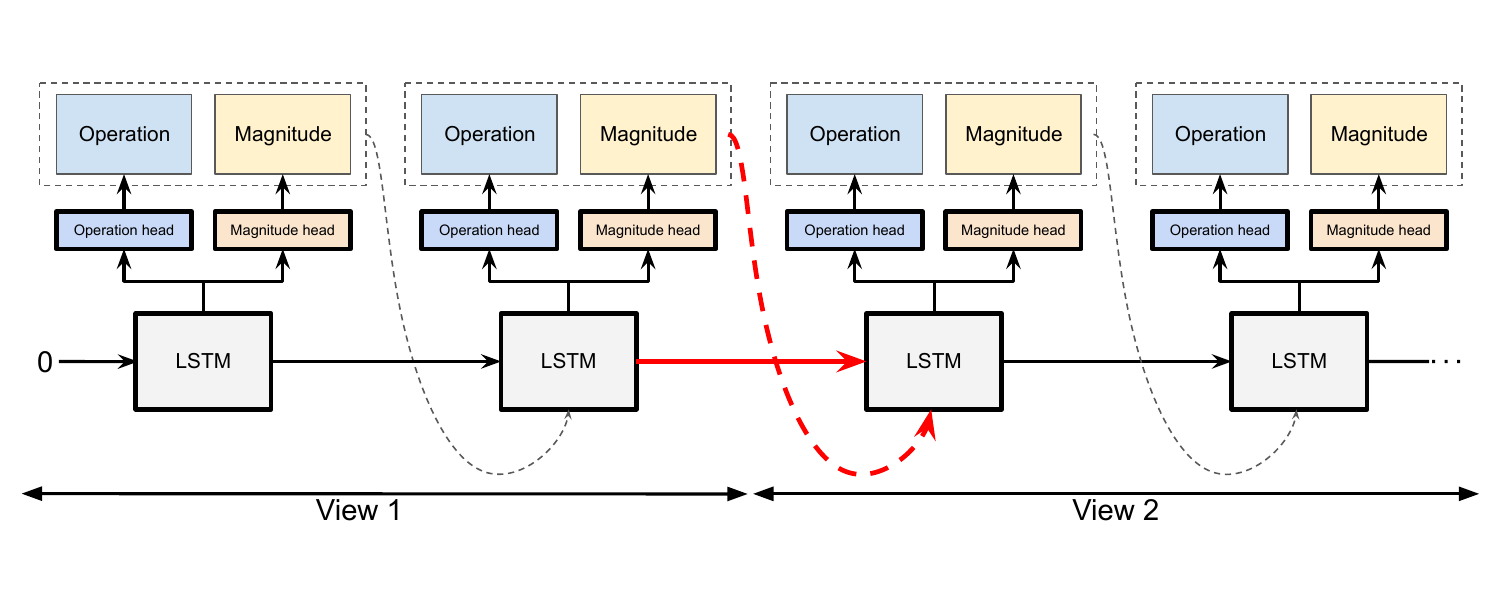}
    \caption{The policy network begins by predicting $N_{\tau}=2$ operations and their corresponding magnitudes for the subpolicy of view 1, and then for the subpolicy of view 2. Each prediction (operation, magnitude) is added to an action history, which is then fed into the next time step as input. In CoViews, we pass the action history and context vector at the end of subpolicy 1 to the LSTM unit again (connection shown in red) to generate subpolicy 2. In the case of IndepViews, we don't pass the action history and context vector from subpolicy 1 to predict subpolicy 2 (the connection shown in red is removed); instead, we start with a new empty action history and a new context vector.}
    \label{fig:ppo_model}
\end{figure}

In order to identify reasonably challenging data augmentation policies for the model during the training, we employ a strategy where, after every K epochs, we learn a new augmentation policy to maximize our Bounded InfoNCE reward function (see \ref{section:bounded_infonce}). 
Our approach formulates the task of finding optimal adaptive augmentation policies as a discrete search problem. 
As shown in Fig. \ref{fig:ppo_model} and inspired by \cite{autoaugment}, our learned policy is represented by a Recurrent Neural Network (RNN) from which we sample the subpolicies. Each subpolicy comprises $N_{\tau}$ transformations to be applied sequentially which are characterized by their operations and magnitudes. For simplicity, we set the probability of each transformation to a fixed probability $p=0.8$ and include the identity transformation in the search space. Additionally, we discretize the range of magnitudes into 11 uniform values.
After training, this policy network will serve as a distribution from which we sample optimal subpolicies. 
The output of the policy network would be of the format $(\text{subpolicy}^{(1)}, \text{subpolicy}^{(2)})$ representing the two subpolicies to apply on the input image to generate the two views, with
$\text{subpolicy}^{(i)} = \{(\text{operation}^{(i)}_{n}, \text{magnitude}^{(i)}_{n}) : n=1,...,N_{\tau} \}$.
Unlike \cite{autoaugment}, we do not fix the number of subpolicies in our policy; instead, we sample new subpolicies from the policy network every time we need to generate views of an input image, thereby increasing the diversity and the likelihood of sampling subpolicies that maximize the reward objective.

Specifically, our network architecture consists of a one-layer LSTM \cite{lstm} with two prediction heads: one for predicting the transformation operation and the other for predicting its magnitude. We repeat this process $2\times N_{\tau}$ times to sample a subpolicy of $N_{\tau}$ transformations for each view. To train such a network, we use Proximal Policy Optimization (PPO) \cite{ppo} with an entropy penalty to encourage the diversity of the generated subpolicies. 
(see Appendix \ref{appendix:ppo} for more details about the search algorithm and a Python-like pseudo code of the trajectory collection phase of PPO).
In the framework of this solution, we present two versions of the policy network:

\paragraph{Cooperative View Policy (CoViews):} We ensure to learn a dependency between the subpolicies by conditioning the generation of the subpolicy for the second view on the subpolicy generated for the first view. This is achievable by passing the subpolicy of the first view, along with the context vector, through the LSTM unit again to generate the subpolicy of the second view. The goal of this method is not only to generate subpolicies that generally maximize the reward objective but rather to also create a compromise between the subpolicies generated for each view for a more efficient and coherent augmentation policy.
In this method, the subpolicies $(\text{subpolicy}^{(1)}, \text{subpolicy}^{(2)})$ generated by the policy network are dependent: hence the naming "cooperative".

\paragraph{Independent View Policy (IndepViews):} We independently and separately generate subpolicies for the first and second views. In this method, the subpolicies $(\text{subpolicy}^{(1)}, \text{subpolicy}^{(2)})$ generated by the policy network are independent and don't share any information (neither action history nor context vector). This is equivalent to a policy network that generates only one subpolicy at a time, and is called twice, once for each view. We preferred maintaining a policy network that outputs two subpolicies in IndepViews - even if they are independent - to keep a similar implementation to CoViews.

\subsection{Bounded InfoNCE Reward}
\label{section:bounded_infonce}
\begin{figure}[ht]
    \centering
    \includegraphics[width=0.5\linewidth]{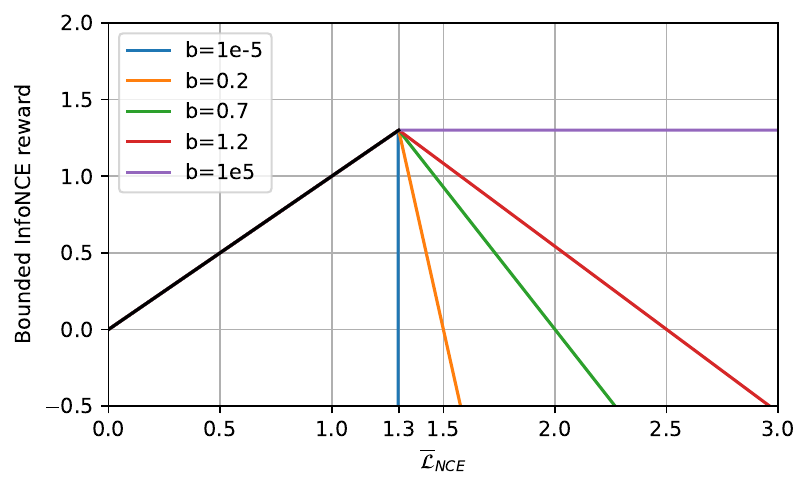}
    \caption{Comparison of Bounded InfoNCE rewards for varying tolerance $b$ values, while keeping the threshold constant at $th=1.3$. A near-zero tolerance aggressively penalizes subpolicies exceeding $th$, while a very large tolerance provides a constant reward equal to $th$ for surpassing subpolicies.}
    \label{fig:bounded_infoNCE_reward}
\end{figure}

Learning augmentations through our previous formulation \ref{eq:min_max_formulation} can be seen as searching for an augmentation policy that maximizes a reward function. 
In this section, we introduce a novel simplistic reward function to carefully lead for reasonably challenging subpolicies that verify conditions (1) and (2). 
The goal is to identify subpolicies that are challenging enough to the current encoder capabilities.
Specifically, we employ a clipping strategy to bound the maximum achievable InfoNCE loss, thereby preventing overly aggressive transformations. To achieve this, we set a dynamic upper-bound that depends on the average achievable InfoNCE loss from the last contrastive training epoch, notably equal to $th$ times $\text{this average loss}$  with $th > 1$. 
Subpolicies exceeding this upper bound are then penalized with linearly decreasing rewards to discourage the policy network from generating them (see Fig. \ref{fig:bounded_infoNCE_reward}). This reward function evaluates subpolicies by calculating the InfoNCE loss of their respective generated views in a batch of views, normalizing it using the average loss from the last epoch, and then returning a reward. Formally, this can be expressed as:

\begin{equation}
\mathcal{R}_{\mathrm{bound}}(\mathcal{L}_{NCE}) = \begin{cases}
    \mathcal{\overline{L}}_{NCE} & \text{if } \mathcal{\overline{L}}_{NCE} < th \\
    \\
    -\frac{th}{b} ( \mathcal{\overline{L}}_{NCE} - (th + b) ) & \text{otherwise}
\end{cases}
\label{eq:bounded_infoNCE_reward}
\end{equation}

where $\mathcal{\overline{L}}_{NCE}$ represents $\mathcal{L}_{NCE}$ normalized by the average InfoNCE loss of the last contrastive training epoch, and $b$ represents a tolerance hyperparameter: a higher value of $b$ implies less penalty for augmentations exceeding the threshold.

\section{Experiments}
\label{section:experiments}

We evaluate the effectiveness of our method in learning more discriminative representations through standard linear evaluation across five vision datasets: CIFAR-10 \cite{CIFAR10}, CIFAR-100 \cite{CIFAR100}, SVHN \cite{svhn}, STL10 \cite{stl10}, and TinyImagenet \cite{tinyImagenet}.
Furthermore, we perform an in-depth analysis of the learned augmentation policies across multiple datasets during training.
See Appendix \ref{appendix:moco} for results of using MoCo \cite{moco} for large negative batch size, and Appendix \ref{appendix:hyperparameters} for full setup details.  

\subsection{Setup}
\label{section:experiments_setup}

In this section, we talk about the setup of our experiments.
All experiments employ SimCLR \cite{simclr} with a temperature of 0.5 and use ResNet-50 as the backbone. We use the SGD optimizer with a base learning rate of $0.03 \times (\text{batch\_size} / 256)$, along with cosine annealing. For each dataset, SimCLR experiments are conducted using six data augmentation strategies:
\begin{enumerate}[label=(\arabic*)]
    \item Random augmentations with random magnitudes.
    \item RandAugment with a fixed magnitude M in \{9, 15, 27\} to compare against randomly selected transformations within a restricted range of magnitudes.
    \item Independent View Policy (IndepViews).
    \item Cooperative View Policy (CoViews).
\end{enumerate}
We use the same data augmentation functions as \cite{autoaugment} (see Appendix \ref{appendix:moco} for more details).
For the bounded InfoNCE reward, we use a threshold of $1.3$ and a tolerance value of $0.2$ 
(see Appendix \ref{appendix:ablation} for an ablation study of these parameters). 
After pretraining our models, we train a linear classifier for linear evaluation for 100 epochs using only random cropping.
All the experiments were conducted on one A100 GPU.

Given the variations in image sizes, for CIFAR-10, CIFAR-100, and SVHN (32x32), we adapted the architecture by replacing the initial 7x7 Convolutional layer with a 3x3 Convolutional layer using stride 1, and we omitted the initial max pooling operation \cite{simclr}, pretraining for 800 epochs on CIFAR-10 and CIFAR-100, and for 400 epochs on SVHN, with a batch size of 512 (1024 negative views). For STL10 (96x96) and TinyImagenet (64x64), we employed a standard ResNet-50 backbone without modifications, pretraining for 400 epochs with a batch size of 256 (510 negative views).

\subsection{Linear Evaluation}
\label{section:linear_evaluation}

\begin{table}[t]
\caption{Top-1 linear probe accuracy on CIFAR-10, CIFAR-100, SVHN, STL10 and TinyImagenet. \ac{IndepViews} and \ac{CoViews}  constantly outperform baseline solutions. Standard deviations are from 5 different random initializations for the linear head. The deviations are small because the linear probe is robust to the random seed.}
\label{fig:linear-evaluation-results}
\centering
\resizebox{1\textwidth}{!}{
\begin{tblr}
{
  row{1} = {c},
  hline{1-2,6,8} = {-}{},
}
                & CIFAR-10          & CIFAR-100        & SVHN             & STL10            & TinyImagenet      \\
RandAug M=9     & 92.61 $\pm$ 0.02  & 68.54 $\pm$ 0.02 & 94.27 $\pm$ 0.00 & 89.98 $\pm$ 0.08 & 30.41 $\pm$ 0.07  \\
RandAug M=15    & 93.16 $\pm$ 0.03  & 70.46 $\pm$ 0.04 & 95.48 $\pm$ 0.02 & 91.05 $\pm$ 0.11 & 31.20 $\pm$ 0.06  \\
RandAug M=27    & 92.55 $\pm$ 0.07  & 69.11 $\pm$ 0.04 & 94.01 $\pm$ 0.01 & 90.88 $\pm$ 0.07 & 30.43 $\pm$ 0.03  \\
Random          & 92.92 $\pm$ 0.04  & 69.56 $\pm$ 0.02 & 96.52 $\pm$ 0.01 & 91.90 $\pm$ 0.03 & 30.23 $\pm$ 0.06  \\
IndepViews (Ours)& 93.68 $\pm$ 0.04 & 72.14 $\pm$ 0.07 & 96.58 $\pm$ 0.03 & 93.01 $\pm$ 0.08 & 35.07 $\pm$ 0.08  \\
CoViews (Ours)   & \textbf{93.79 $\pm$ 0.08} & \textbf{72.28 $\pm$ 0.13} & \textbf{96.69 $\pm$ 0.07} & \textbf{93.67 $\pm$ 0.05} & \textbf{36.29 $\pm$ 0.08} \\
\end{tblr}
}
\end{table}

The linear evaluation results shown in Table \ref{fig:linear-evaluation-results} demonstrate a consistent improvement in linear probe accuracy of both \ac{IndepViews} and \ac{CoViews} over SimCLR with both RandAugment and random augmentations, indicating that simply finding an optimal magnitude for all transformations or randomly applying transformations with random magnitudes is not sufficient, and more sophisticated subpolicies learned by IndepViews and CoViews are required to learn rich representations. 
Moreover, we observe that CoViews outperforms IndepViews. This suggests that learning and training with dependent cooperative subpolicies lead to the generation of better pairs, ultimately resulting in enhanced learned representations.
It's also worth noting that RandAugment with a magnitude of M=15 consistently outperforms both M=9 and M=27, indicating that augmentations that are either too easy or too hard result in worse representations, and a middle-ground is preferable.

\begin{wraptable}{r}{0.5\textwidth}
\footnotesize	

\begin{tblr}{
  column{2} = {c},
  column{3} = {c},
  hline{1-2,5} = {-}{},
  0.25\textwidth,
}
       & \textbf{$\text{CIFAR-10}^{10\%}$} & \textbf{$\text{STL-10}^{5\%}$}   \\
Random & 69.20 $\pm$ 0.12          & 61.68 $\pm$ 0.04         \\
IndepViews (Ours)   & 74.48 $\pm$ 0.07          & \textbf{71.08 $\pm$ 0.15}\\
CoViews (Ours)   & \textbf{75.05 $\pm$ 0.08} & 69.73 $\pm$ 0.13       
\end{tblr}
\caption{linear probe accuracies on $\text{CIFAR-10}^{10\%}$ and $\text{STL-10}^{5\%}$ datasets. Standard deviations are from 5 different random seeds for the linear head.}
\label{table:less_data}
\end{wraptable}

Additionally, we tested our method on CIFAR-10 with only 10\% of the training images (5000 images) and STL-10 with only 5\% of the training images (5000 images) for pretraining. We refer to these as $\text{CIFAR-10}^{10\%}$ and $\text{STL-10}^{5\%}$. Results shown in Table \ref{table:less_data} indicate that the model shows an increase in performance by 5.28\% and 5.85\% for IndepViews and CoViews respectively in CIFAR-10, and a 9.4\% and 8.05\% improvement for IndepViews and CoViews respectively in STL-10. Both methods show a smaller improvement over using random augmentation when trained on the full dataset (see Table \ref{fig:linear-evaluation-results}). This suggests that reasonably challenging transformations can compensate for the lack of pretraining images.

\subsection{Inspecting the learned augmentation policies}

We inspect the evolution of the probability of each transformation in the policy through epochs, for both \ac{IndepViews} and \ac{CoViews}.

\begin{figure}[ht]
    \centering
    \includegraphics[width=1\linewidth]{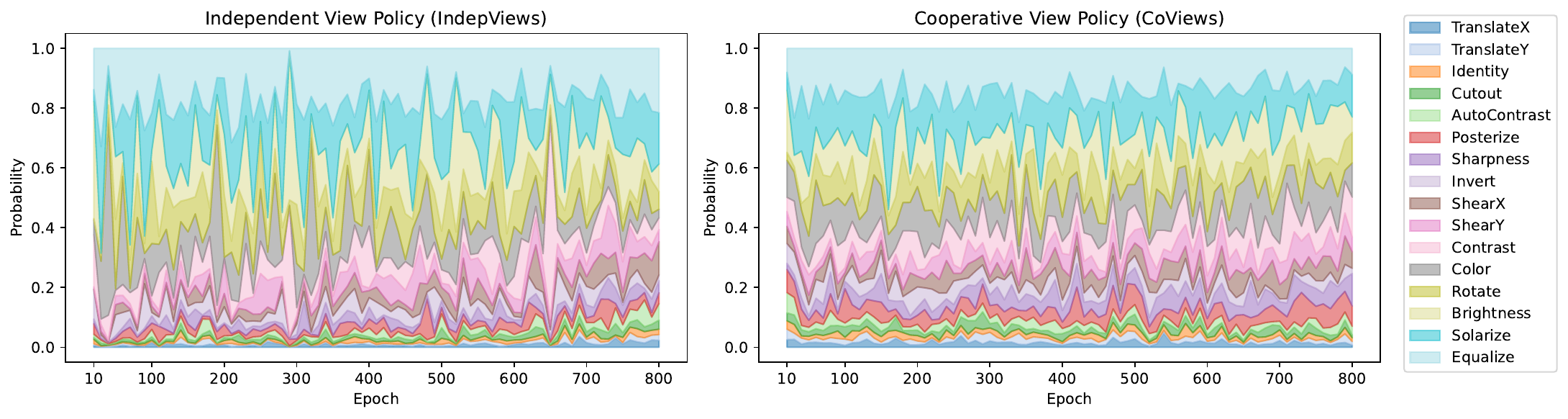}
    \caption{Comparison of the evolution of transformation probability in the learned adaptive augmentation policies between \ac{IndepViews} and \ac{CoViews} on CIFAR-10 dataset. }
    \label{fig:IAAD_DAAD}
\end{figure}

We notice in Fig. \ref{fig:IAAD_DAAD} that IndepViews exhibits a highly variable probability distribution with frequent spikes, particularly focusing on highly distorting transformations like Equalize, Solarize, and Brightness. However, this variability and noise in the probability distribution might indicate potential instability in the learning process, which could impact its effectiveness. Despite initially giving very low probabilities for some transformations such as translation and auto contrast, we can notice an increase in their probabilities by the end of the training. 
In contrast, CoViews demonstrates a more consistent and balanced distribution of probabilities. While still displaying minor fluctuations, it avoids the pronounced spikes observed in IndepViews, indicating a smoother learning process. Although transformations like Equalize, Solarize, and Brightness receive slightly higher probabilities due to their high effect on distorting the images, there is a more equitable distribution across all transformations. This stability in the probability distribution potentially makes this method better for learning adaptive augmentation policies.

\begin{figure}[ht]
    \centering
    \includegraphics[width=0.8\linewidth]{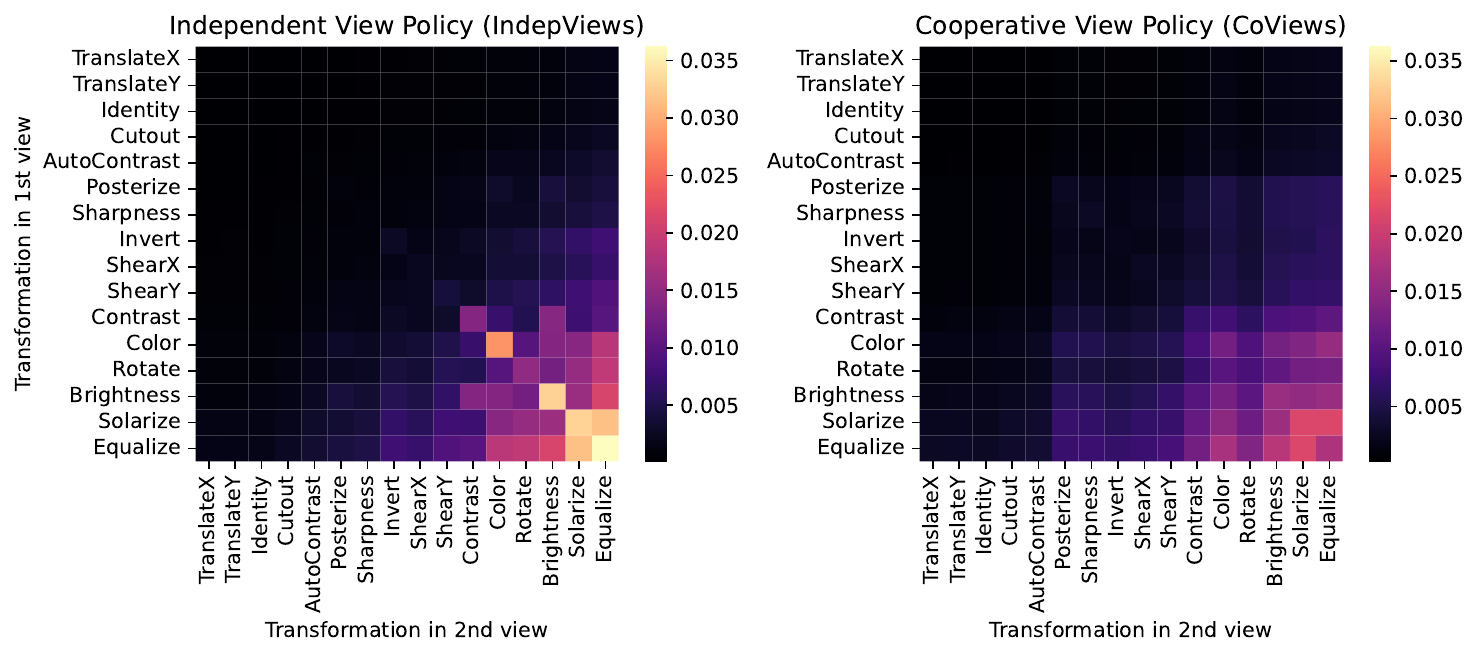}
    \caption{A comparison between the co-occurrence matrix of transformations in view 1 and view 2 of both \ac{IndepViews} and \ac{CoViews}. Each value in the matrix represents the frequency of the co-occurrence of corresponding transformations across the two views.}
    \label{fig:co-occurrence}
\end{figure}

We also inspect the co-occurrence of transformations within the subpolicies generated for each view in Fig. \ref{fig:co-occurrence}. We observe that IndepViews tends to prioritize transformations such as Equalize, Solarize, and Brightness, often applying the same transformation in generating both views. This is because IndepViews needs to ensure that the views are sufficiently challenging without any information about the transformation applied for the other view, thus playing it safe and focusing on highly distorting transformations.
On the other hand, CoViews demonstrates a more balanced and diverse utilization of transformations, allowing for various combinations of subpolicies. This highlights that we do not necessarily need to heavily rely on highly distorting transformations to create challenging positive pairs. Instead, cooperative view generation can effectively achieve this goal.

\section{Discussion}
\paragraph{Independent View Policy vs Cooperative View Policy} 
Independent View Policy and Cooperative View Policy yield comparable results and both surpass baseline solutions. This proves the effectiveness of the foundational components of both variants: our search algorithm and the bounded InfoNCE reward function in learning robust adaptive augmentation policies. However, linear evaluation reveals that CoViews shows superior performance to IndepViews, due to its diverse and stable learned policies. Hence, we favor this approach despite its more complex implementation. But for the sake of simplicity, individuals wishing to use our solution may opt for the Independent View Policy method, as it is easier to implement with a slight decrease in performance.

\paragraph{Why using a neural network with PPO as a search algorithm ?} 
In our implementation, we chose to utilize a neural network and train it with PPO to generate policies, despite the emergence of faster and more efficient search algorithms for learning data augmentations in recent research \cite{randaugment, faa}. While it's easy to switch to such methods for \ac{IndepViews}, integrating them for CoViews is less straightforward. Recurrent Neural Networks present an intuitive solution that enables learning a dependency relationship by recursively passing the subpolicy of view 1 along with the context vector to generate the subpolicy of view 2.

\paragraph{Computational efficiency}
\label{section:overhead}
We consider one of the primary strengths of our solution to be its minimal computational overhead as shown in table \ref{tab:execution-time}. Unlike prior approaches that involve grid searching and training multiple models to determine optimal augmentations, which can be highly computationally intensive and often impractical, our method offers a more efficient alternative. Our solution shows a computational overhead of 1.57 in our smallest dataset and 0.04 in our biggest dataset which is negligible.
Furthermore, our solution can be accelerated further by adjusting parameters such as increasing K the number of epochs to generate a new policy, reducing the number of PPO epochs, or reducing the number of collected subpolicies (trajectories) in each PPO epoch.

\begin{table}[t]
\centering
\caption{The number of GPU hours required for pretraining on the benchmark datasets for both random augmentation (baseline) and using our methods IndepViews and CoViews. We use one A100 GPU.}
\begin{tabular}{lcccccc} 
\hline
          & CIFAR10 & CIFAR100 & SVHN   & STL10 & STL10+unlabeled & TinyImagenet  \\ 
\hline
Baseline  & 21      & 19.8     & 15.8   & 3.5   & 35.7            & 34.7          \\
Ours      & 30.6    & 27.9     & 22.3   & 9.0   & 40.2            & 36.2          \\
Overhead  & 0.46    & 0.41     & 0.41   & 1.57  & 0.13            & 0.04          \\
\hline
\end{tabular}
\label{tab:execution-time}
\end{table}

\paragraph{Limitation} 
\label{section:limitation}
We believe that our solution could benefit from integrating more efficient and suitable search algorithms than Reinforcement Learning. Moreover, our solution requires transformation functions to be first implemented by experts and cannot come up with new transformations. This may limit its applicability to new data modalities and could be constrained by the efficiency of the manually crafted transformation functions. Additionally, although we believe that the simplicity of the InfoNCE Bounded Reward function is an advantage of our method, developing more complex and sophisticated evaluation functions might be beneficial. Furthermore, we would like to test our method on other data modalities such as text and graphs in future works.

\section{Conclusion}

In this work we develop a new approach to learning optimal adaptive augmentations during the training to improve contrastive learning, we achieve this by learning a new augmentation policy that maximizes our Bound InfoNCE reward after each K epoch of contrastive training. We introduce two solutions: IndepViews, which learns one common policy to generate both views, and CoViews which learns two different policies that are conditioned on each other to create a compromise between the generated views. Through linear evaluation, we show that our solution improves over baseline solutions with minimal overhead and that training with cooperative views yields better performance. In conclusion, our research demonstrates the feasibility and benefits of rapidly learning adaptive augmentation policies to generate dependent cooperative views, thereby enhancing contrastive learning, and paving the way for more efficient self-supervised learning.

\newpage
\bibliographystyle{plain} 
\bibliography{neurips_2024.bib}

\newpage

\newpage
\appendix

\section{Notation and definitions}

\begin{table}[h]
\centering
\begin{tblr}{} 
\hline
\textbf{Notation}   & \textbf{Definition} \\ \hline
$N$                 & Batch size \\ \hline
$X_i$               & View i of batch X \\ \hline
$X_{i,j}$           & Image j of View i of batch X \\ \hline
$A$                 & Set of data augmentations \\ \hline
$N_{\tau}$          & The number of transformations in the subpolicy \\  \hline
$K$                 & The number of epochs elapsed before learning a new data augmentation policy \\ \hline
$Q$                 & The number of most recent policies to use for contrastive training \\ \hline
$th$                & The infoNCE upper bound in R bound \\ \hline
$b$                 & The tolerance parameter in the R bound \\ \hline
$\mathcal{R}_{bound}$   & the Bounded InfoNCE reward function \\ \hline
$\text{subpolicy}^{(i)}$   & Subpolicy $i$ of view $i$ used in \ac{CoViews} and \ac{IndepViews} \\ \hline
$\mathcal{L}_{NCE}$ & InfoNCE loss \\ \hline
$\mathcal{\overline{L}}_{NCE}$  & InfoNCE loss normalized by the average InfoNCE loss of the last epoch \\ \hline

\end{tblr}
\end{table}
\section{Augmentation transformation details}
We use the same data augmentation transformations set as \cite{autoaugment}. Each transformation has a range of magnitude except for binary transformations that behave the same regardless of predicted magnitude.

\begin{table}[ht]
\centering
\caption{Transformtions along with their magnitude range.}
\begin{tblr}{
    hline{1-2,18} = {-}{},
    colspec = {Q[0.15,l] Q[0.7,l] Q[0.15,c]},
}
\textbf{Name}         & \textbf{Description} & \textbf{Min, Max} \\
ShearX       & shear the image along the horizontal axis with magnitude & -0.3, 0.3           \\
ShearY       & shear the image along the vertical axis with magnitude & -0.3, 0.3           \\
TranslateX   & translate the image in the horizontal axis with magnitude & -0.45, 0.45          \\
TranslateY   & translate the image in the vertical axis with magnitude & -0.45, 0.45          \\
Rotate       & rotate the image with magnitude degrees & -30, 30            \\
AutoContrast & automatically adjusts the contrast of the image & -             \\
Invert       & invert the pixels of the image & -             \\
Equalize     & equalize the image histogram & -             \\
Solarize     & invert the pixels above a magnitude threshold & 0, 256           \\
Posterize    & reduce the number of bits for each color to magnitude & 4, 8            \\
Contrast     & adjust image contrast with magnitude 0 is grey and magnitude 1 is original image & 0.1, 1.9           \\
Color        & adjust the color of the image such that magnitude 0 is black and white and magnitude 1 is the original image   & 0.1, 1.9           \\
Brightness   & brightness adjustment such that magnitude 0 is the black image and 1 is the original image & 0.1, 1.9           \\
Sharpness    & magnitude 0 is a blurred image and 1 is the original image & 0.1, 1.9           \\
Cutout       & cutout a random square from the image with a side length equal to the magnitude percentage of pixels & 0, 0.2           \\
Identity     & Identity function, no transformation & -           
\end{tblr}
\end{table}

\section{Proximal Policy Optimization hyperparamteres}
\label{appendix:ppo}
Since we work with non-differentiable transformations, we used Proximal Policy Optimization to train our policy network. All implementation details for PPO are presented in Table \ref{table:hyperparameters}. Generally, the PPO algorithm consists of two phases: (1) collecting trajectories from the environments and (2) updating the policy using the collected trajectories. In the context of our solution, we can consider that our environment has an episode length equal to one. This is because we can construct the subpolicy in only one inference and then proceed to generate another subpolicy. Therefore, collecting trajectories becomes equivalent to collecting subpolicies for a batch of images. 
A Python pseudo-code of the collection phase is available at Listing \ref{algo:collect-trajectories}.
As we are dealing with only one-step environments, we do not train a critic, and thus, we do not employ Generalized Advantage Estimation. Instead, we utilize normalized rewards.

\begin{lstlisting}[language=Python, caption={Python pseudo-code for collecting samples (trajectories) for the PPO algorithm to update the policy network}, label={algo:collect-trajectories}]
# N          : The number of samples to collect for a PPO iteration
# bs         : The batch size 
# dataset    : The dataset
# policy_net : The policy network that generates the policy
# InfoNCE    : InfoNCE criterion function
# Reward     : Bounded InfoNCE reward function
# encoder    : Encoder network

actions, log_ps, rewards = [], [], [] 

for _ in range(num // bs):
    # Sample `bs` images and `bs` subpolicies 
    img = dataset.sample(bs)
    action, log_p = policy_network.sample(bs)

    # Apply subpolicies to generate the views `x1` and `x2` 
    sub_policy1, sub_policy2 = action
    x1 = apply(img, sub_policy1)
    x2 = apply(img, sub_policy2)

    # Get the representations `z1` and `z2` of the views 
    z1 = encoder(x1)
    z2 = encoder(x2)

    # Calculate the reward
    loss = InfoNCE(z1, z2)
    reward = Reward(loss)

    # Store the action, its log probability, and its reward
    actions.append(action)
    log_ps.append(log_p)
    rewards.append(reward)

return actions, log_ps, rewards

\end{lstlisting}

\newpage
\section{Detailed training hyperparameters}
\label{appendix:hyperparameters}
The following table contains the hyperparameters used in SimCLR pertaining, linear probe evaluation and PPO model to train the policy networks.
\begin{table}[h]
\centering
\caption{Hyperparameters used in our experimental setup.}
\begin{tblr}{
  hline{1-2,11-12,19-20,31} = {-}{},
}
\textbf{SimCLR Params~}                   & \textbf{Value}\\
Feature dim                               & 2048\\
Temperature                               & 0.5\\
Batch size                                & {512 (Cifar10 / Cifar100 / SVHN)\\256 (STL10 / TinyImagenet)}\\
Learning rate                             & {0.06 (Cifar10 / Cifar100 / SVHN)\\0.03 (STL10 / TinyImagenet)} \\
Schedule                                  & Cos annealing\\
Schedule - warmup epochs                  & 10\\
Momentum                                  & 0.9\\
Weight decay                              & 5e-4\\
Epochs                                    & {800  (Cifar10 / Cifar100)\\400 (SVHN / STL10 / TinyImagenet)}\\
\textbf{Linear Classifier Params}         &\\
Batch size                                & 256\\
Learning rate                             & 30\\
Schedule                                  & Cosine annealing\\
Schedule - warmup epochs                  & 0\\
Momentum                                  & 0.9\\
Weight decay                              & 0.0\\
Epochs                                    & 100                                 \\
\textbf{PPO Parameter}                    & \textbf{Value}                      \\
Optimizer                                 & Adam                                \\
Learning rate                             & 5e-5                                \\
PPO epochs                                & 100                                 \\
Number of collected subpolicies per epoch & 128                                 \\
Number of policy updates per epoch        & 4                                   \\
Policy update batch size                  & 16                                  \\
Entropy coefficient                       & 0.05                                \\
Clip parameter                            & 0.2                                 \\
Grad clip                                 & None                                \\
Use critic                                & No                                  \\
Advantage                                 & Use normalized reward               \\
\end{tblr}
\label{table:hyperparameters}
\end{table}

\section{Policy Probabilty}
\label{appendix:proba}
To ensure more stable training, we aim for a smooth transition to newer policies by maintaining a history of the $Q$ most recently generated policies. We achieved this by assigning exponentially decreasing probabilities $(p_1 > ... > p_Q)$ to the most recent policies with $p_1$ being the probability of the most recent policy. We draw inspiration from a geometric probability distribution, but since we have only Q policies we need to adapt it by putting $p_i = \frac{p(1-p)^{i-1}}{M}$ with $M$ being a normalizing factor to assure that $\sum_{i=1}^Qp_i=1$:

\begin{equation}
\begin{split}
\sum_{i=1}^Qp_i=1 & \Rightarrow \sum_{i=1}^Q  \frac{p(1-p)^{i-1}}{M}  =1 \\
                  & \Rightarrow \frac{p}{M} \cdot \sum_{i=1}^Q (1-p)^{i-1}  =1 \\
                  & \Rightarrow \frac{p}{M} \cdot \frac{1-(1-p)^{i-1}}{1-(1-p)}  =1 \\
                  & \Rightarrow \frac{1-(1-p)^{i-1}}{M}  =1 \\
                  & \Rightarrow M=1-(1-p)^{i-1} \\
                  & \Rightarrow p_i=\frac{p(1-p)^{i-1}}{1-(1-p)^{i-1}} \\
\end{split}
\end{equation}

To test the effect of the policy queue, we conducted experiments without utilizing it, and some of these experiments failed, particularly when employing IndepViews with high thresholds, because the model can learn a degenerate augmentation policy, resulting in the failure of the entire training process. However, by implementing a queue of policies, we can still train the model using previous policies, even if the most recent policy is degenerate, thus saving the model from failing until learning the next new policy.

\section{Dynamic Adaptive Augmentation (IndepViews / CoViews) Pseudo Algorithm}
\label{pseudo-algorithm}
The following represents a pseudo-algorithm of our algorithm. We start with a "warmup phase" where the model is trained using random augmentation to establish a baseline performance. Then, for each K epoch, we learn a new policy either using IndepViews or CoViews and add it to the queue of policies, and we train the model using these policies until the end of the training.

\begin{algorithm}
\caption{Dynamic Adaptive Augmentation}
\begin{algorithmic}[1]
    \State \textbf{function} learn\_new\_policy($f$) \Comment{Learn new policy for $f$}
    \State \textbf{function} contrastive\_train($f$, augmentation\_policy) \Comment{Train $f$ using 'augmentation\_policy'}
    \\
    \State adaptive\_policies $\gets$ queue(size=Q)
    \\
    \For{$\text{epoch} = 1$ to $\text{num\_epochs}$}
        \If{$\text{epoch} > \text{warmup\_epochs}$ \textbf{and} $\text{epoch \% } K = 0$} \Comment{adaptive policy generation phase}
            \State $\text{policy} \gets \text{learn\_new\_policy}(f)$
            \State $\text{adaptive\_policies.push}(\text{policy})$
        \EndIf
        \\
        \If{$\text{epoch} \leq \text{warmup\_epochs}$}
            \State $f \gets \text{contrastive\_train}(f, \text{random\_policy})$ \Comment{warmup phase}
        \Else
            \State $f \gets \text{contrastive\_train}(f, \text{adaptive\_policies})$ \Comment{contrastive training phase}
        \EndIf
    \EndFor
\end{algorithmic}
\end{algorithm}

\section{More experiments}

In this section we present additional experiments that we performed, notably 
an ablation study to study the effect of the parameters of the Bounded InfoNCE reward function (see \ref{section:bounded_infonce}), 
experiments using the SimCLR original augmentations \cite{simclr} instead of the augmentation used in AutoAugment \cite{autoaugment}.
and experiments using MoCo \cite{moco} for larger batch sizes.

\subsection{Bounded InfoNCE reward ablation study}
\label{appendix:ablation}
In this section we perform an ablation study on the parameters of the Bounded InfoNCE Reward function: the threshold parameter and the tolerance parameter.

\subsubsection{Threshold parameter}
\begin{figure}[ht]
    \centering
    \includegraphics[width=\linewidth]{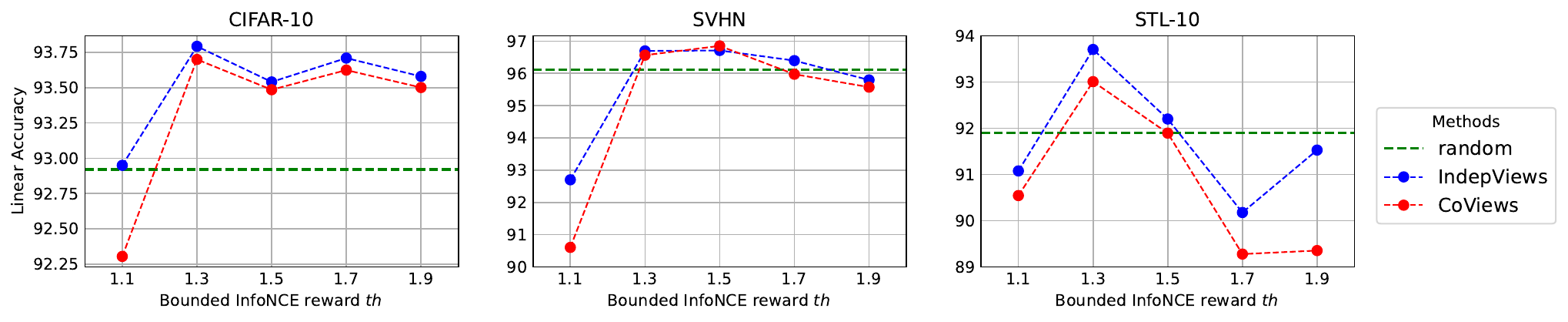}
    \caption{Linear evaluation over CIFAR-10, SVHN, and STL10 using a threshold value in the range [1.1, 1.3, 1.5, 1.7, 1.9]. All experiments use a tolerance value of 0.2.}
    \label{th_ablation}
\end{figure}

To test the effect of the threshold parameter of the Bounded InfoNCE reward function, we pretrained multiple models using thresholds in the range of [1.1, 1.3, 1.5, 1.7, 1.9] on different datasets: CIFAR-10, SVHN, and STL-10, as illustrated in Fig. \ref{th_ablation}. We observed that some datasets are more sensitive to the threshold parameters than others (STL-10 is more sensitive than CIFAR-10 and SVHN). All datasets exhibit lower performance when the threshold is set to its lowest value (th=1.1) due to less challenging learned augmentation policies for the model that led to very poor representations. Conversely, higher threshold values generally result in decreased performance because the learned augmentation policies become excessively challenging, often producing views with very little task-relevant information that are difficult to differentiate. Notably, we observed a consistent transferability of the optimal threshold parameter across datasets, suggesting that the same threshold values that worked well in experimental datasets can be applied without always needing to search for new optimal values. From the results depicted in Fig. \ref{th_ablation}, it's shown that optimal performance is achieved with a threshold value around 1.3.

\subsubsection{Tolerance parameter}

To test the effect of the tolerance parameter of the Bounded InfoNCE reward function, we pretrained multiple models using extreme values for the tolerance parameters:
\begin{enumerate}[label=(\arabic*)]
    \item Very small tolerance value ($b=$1e-5): this implies that subpolicies surpassing the threshold are heavily penalized with very large penalties. (see Fig. \ref{fig:bounded_infoNCE_reward})
    \item A very large tolerance value ($b=$1e5): this implies that subpolicies surpassing the threshold are not penalized at all and will receive a reward equal to the threshold itself. (see Fig. \ref{fig:bounded_infoNCE_reward})
\end{enumerate}

\begin{figure}[ht]
    \centering
    \includegraphics[width=1\linewidth]{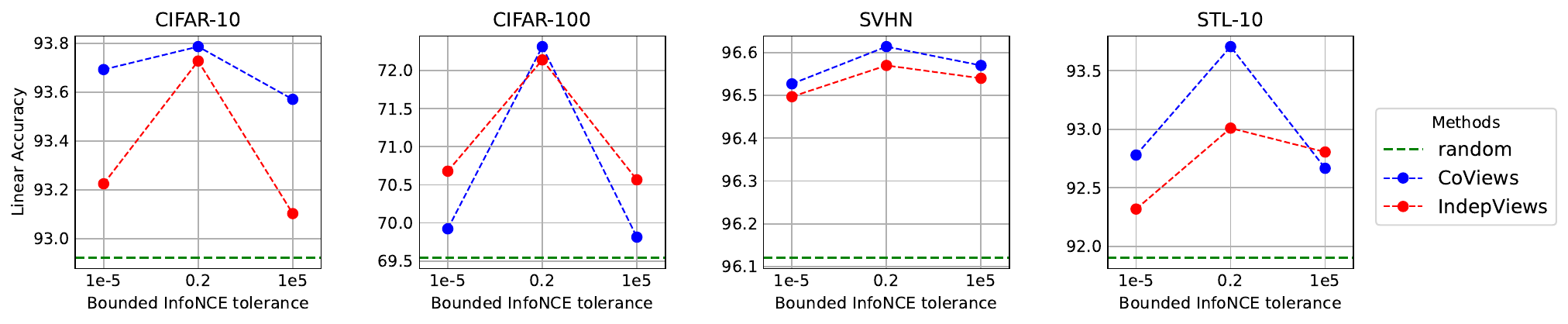}
    \caption{Linear evaluation over CIFAR-10, CIFAR-100, SVHN, and STL10 using a tolerance value in the range [1e-5, 0.2, 1e5]. All experiments use a threshold value of 1.3.}
    \label{fig:tolerance_ablation}
\end{figure}

We compare both these setups with the use of a tolerance value of 0.2, which reasonably penalizes subpolicies surpassing the threshold, as employed in all of our experiments. As shown in Fig. \ref{fig:tolerance_ablation}, we can notice that heavily penalizing ($b=$1e-5) or not penalizing subpolicies ($b=$1e5) results in a decrease in performance for all datasets. This is because:
\begin{enumerate}[label=(\arabic*)]
    \item Heavy aggressive penalties will make the policy network avoid subpolicies giving an InfoNCE loss close to the threshold, as slightly surpassing the threshold will still result in a very big penalty. Therefore, the policy network will only play it safe and generate subpolicies giving lower InfoNCE loss than the threshold, resulting in easy views.
    \item  Not penalizing and giving rewards equal to the threshold for subpolicies surpassing the threshold will make the model more confident about using very hard augmentations, as it ensures getting a maximum reward, resulting in hard views.
\end{enumerate}

 So it's better to reasonably penalize subpolicies and keep a smooth reward function in order to identify a reasonably challenging policy.

\subsection{SimCLR original augmentation}
We also evaluated our method using the original SimCLR augmentation set - as shown in Table \ref{table:magnitudes} - to compare its performance against the optimal augmentation determined by supervised evaluation. To achieve this, we trained two encoders using the \ac{CoViews} with the same augmentation transformations but using different magnitudes:

\begin{enumerate}
    \item In the first experiment, we utilized the same magnitudes as in SimCLR to explore whether performance could be enhanced even with identical magnitudes.
    \item In the second experiment, we employed larger magnitudes compared to SimCLR to investigate whether a better augmentation policy could be learned using the same transformations.
\end{enumerate}

\begin{table}[t]
\centering
\caption{Both experiments use the same augmentation transformations as SimCLR. Experiment 1 uses the same magnitudes while Experiment 2 uses stronger magnitudes.}
\begin{tabular}{lcccc} 
\hline
             & Brightness & Contrast  & Saturation & Hue           \\ 
\hline
CoViews (same magnitudes) & 0.6 - 1.4  & 0.6 - 1.4 & 0.6 - 1.4  & -0.1 - 0.1        \\
CoViews (bigger magnitudes) 2 & 0.1 - 1.9  & 0.1 - 1.9 & 0.1 - 1.9  & -0.45 - 0.45  \\
\hline
\end{tabular}
\label{table:magnitudes}
\end{table}

The linear evaluation results in Table \ref{table:simclr_aug_results} have shown very comparable performance between the SimCLR baseline and both experiments, with a slight improvement observed with CoViews over the baseline when using the same magnitudes (first experience). The similarity in linear probe accuracy can be due to this set of augmentations having been previously identified as optimal through supervised evaluation \cite{simclr}. Therefore, we do not expect any significant improvements from CoViews. 
This demonstrates the efficiency of our method in identifying state-of-the-art augmentations in a single training run without requiring extensive evaluation.

\begin{table}[t]
\centering
\caption{Linear evaluation accuracy using CoViews with SimCLR augmentations}
\begin{tblr}{
  column{even} = {c},
  column{3} = {c},
  hlines,
}
                  & SimCLR           & CoViews (same magnitudes)     & CoViews (bigger magnitudes)     \\
Linear evaluation & 92.93 $\pm$ 0.04 & \textbf{93.04 $\pm$ 0.01} & 92.91 $\pm$ 0.02 
\end{tblr}
\label{table:simclr_aug_results}
\end{table}

\subsection{Experiments using the MoCo framework}
\label{appendix:moco}

Experiments in the main body of the paper are all based on using SimCLR with a relatively small batch size of up to 512. To test our method on different contrastive learning frameworks and to check its effectiveness with larger batch sizes, we conducted experiments using MoCo \cite{moco} with CIFAR-10 dataset employing batch sizes of 4096 and 65536. We used 10 warmup epochs and pre-trained for a total of 200 epochs. 

\begin{table}[h]
\centering
\caption{This table shows the results of KNN classification accuracy of Pretraining MoCo on CIFAR-10 using random augmentations, \ac{IndepViews}, and \ac{CoViews}.}
\begin{tabular}{lcc} 
\hline
            & \textbf{Queue size = 4096} & \textbf{Queue size = 65536} \\ 
\hline
Random      & 81.13     & 81.07     \\
IndepViews         & 82.23     & 82.27     \\
CoViews         & \textbf{82.98}     & \textbf{82.83}     \\
\hline
\end{tabular}
\label{table:moco}
\end{table}

Results in Table \ref{table:moco} demonstrate that both our methods outperform the baseline solution using random augmentations, with CoViews showing superior performance compared to IndepViews. For both IndepViews and CoViews, we get a computational overhead of 0.26 over the baseline. 
To achieve this in Moco, we just replaced the InfoNCE function in the bounded reward function with the MoCo InfoNCE loss function. However, the original MoCo function is asymmetric, as it treats the query and key representations differently. To simplify, we made it symmetric by calculating an additional loss where the query serves as the key and vice versa. We then feed the mean of both losses to the bounded reward function to train the policy network.

\newpage

\end{document}